\documentclass[letterpaper]{article} 
\usepackage[]{aaai23}  
\usepackage{times}  
\usepackage{helvet}  
\usepackage{courier}  
\usepackage[hyphens]{url}  
\usepackage{graphicx} 
\usepackage{natbib}  
\usepackage{caption} 
\usepackage{algorithm}
\usepackage[noend]{algorithmic}

\usepackage{xspace}

\usepackage{amsfonts, amssymb, amsmath}
\usepackage{color}
\usepackage[inline]{enumitem}
\usepackage{float}

\newcommand{\algName}[1]{\textsf{#1}\xspace}
\newcommand{\mcts}{\textsf{MCTS}\xspace}
\newcommand{\wsts}{\textsf{WSTS}\xspace}
\newcommand{\bs}{\textsf{BS}\xspace}
\newcommand{\rmbs}{\textsf{EM-BS}\xspace}
\newcommand{\ignore}[1]{}

\usepackage{newfloat}
\usepackage{listings}
\DeclareCaptionStyle{ruled}{labelfont=normalfont,labelsep=colon,strut=off} 

\floatstyle{ruled}
\newfloat{listing}{tb}{lst}{}
\floatname{listing}{Listing}
\title{Wall Street Tree Search: Risk-Aware Planning for Offline Reinforcement Learning}

\author {
    Dan Elbaz, \textsuperscript{\rm 1, 2}
    Gal Novik,  \textsuperscript{\rm 1}
    Oren Salzman \textsuperscript{\rm 2}
}
\affiliations {
    \textsuperscript{\rm 1} Intel Labs\\
    \textsuperscript{\rm 2} Technion Israel Institute of Technology
Haifa, Israel\\
    dan.elbaz@intel.com, gal.novik@intel.com, osalzman@cs.technion.ac.il
}

\begin{document}

\maketitle

\begin{abstract}

Offline reinforcement-learning (RL) algorithms learn to make decisions using a given, fixed training dataset without online data collection. 
This problem setting is captivating because it holds the promise of utilizing previously collected datasets without any costly or risky interaction with the environment. However, this promise also bears the drawback of this setting as the restricted dataset induces uncertainty because the agent can encounter unfamiliar sequences of states and actions that the training data did not cover. 
To mitigate the destructive uncertainty effects, we need to balance the aspiration to take reward-maximizing actions with the incurred risk due to incorrect ones.
In financial economics, modern portfolio theory (MPT) is a method that risk-averse investors can use to construct diversified portfolios that maximize their returns without unacceptable levels of risk.
We propose integrating MPT into the agent's decision-making process, presenting a new simple-yet-highly-effective risk-aware planning algorithm for offline RL.
Our algorithm allows us to systematically account for the \emph{estimated quality} of specific actions and their \emph{estimated risk} due to the uncertainty.
We show that our approach can be coupled with the Transformer architecture to yield a state-of-the-art planner, which maximizes the return for offline RL tasks. Moreover,  our algorithm reduces the variance of the results significantly compared to conventional Transformer decoding, which results in a much more stable algorithm---a property that is essential for the offline RL setting, where real-world exploration and failures can be costly or dangerous.

\end{abstract}

\section{Introduction}


Reinforcement Learning (RL) is concerned with an agent learning how to take actions in an environment to maximize the total reward it obtains. The RL agent typically learns by trial and error, involving online interaction with the environment to collect experiences~\citep{r:1}. But learning in the real world may be undesirable, as online data acquisition is often costly, time-consuming, or even dangerous.
Offline RL aims to bridge the gap between RL algorithms and real-world systems by leveraging an existing dataset or batch to learn how to make decisions in an offline stage without any online interactions with the environment~\citep{r:2}.

However, learning solely from offline data is a double-edged sword. On the one hand, it enables applications in domains where online exploration is avoided, e.g., in healthcare~\citep{r:63, r:75, r:76}, autonomous driving ~\citep{r:79}, and recommendation systems~\citep{r:77, r:78}. On the other hand, it poses a major algorithmic challenge as we are faced with high levels of uncertainty.
As in offline RL, we typically want the learned policy to perform better than the policy that collected the data. Consequently, we must execute a different sequence of actions from the sequences stored in the batch.
As a result, the agent encounters unfamiliar state-action sequences,
which induces \emph{subjective uncertainty} (uncertainty that comes from ignorance due to the limited size of the training batch). 
This uncertainty can lead to erroneous value estimations, a phenomenon known as \emph{distributional shift}~\citep{r:23}, which is one of the central challenges of offline RL. Moreover, \emph{objective uncertainty}, which is due to inherent system stochasticity, further increases uncertainty and aggravates the offline RL problem, preventing the agent from learning an optimal policy.

One promising way to solve the offline RL problem is using \emph{model-based RL} (MBRL). MBRL divides the RL problem into two stages:
The first stage is learning an approximate environment model (referred to as the transition or dynamics model) with the data. In the second stage, this model is used for decision-making (i.e., planning or policy search), usually via Model-Predictive Control (MPC)~\citep{r:7}.
An advantage of using MBRL is that we can benefit from a convenient and powerful supervised learning workhorse in the model-learning stage, which allows us to generalize to states outside the support of the batch. However, due to the distribution shift, the model becomes increasingly inaccurate as we get further from the points in the batch.
As a result, a planner that tries to obtain the highest possible expected reward under the model without any precautions against model inaccuracy can result in ``model-exploitation''~\citep{r:2}---a situation where the planner prefers predictions with higher returns than would be obtained from the actual environment, resulting in poor performance.

Recent offline RL approaches suggest drawing from the
tools of large-scale language modeling, using a Transformer architecture~\citep{r:57} to model distributions over trajectories~\citep{r:3, r:4}. 
However, large-scale language models are 
trained on millions of high-quality web pages; hence their decoding methods are not designed to be immune to the effects of objective and subjective uncertainty prevalent in offline RL.
In this work, we aim to address the question of how we can modify these decoding algorithms to tackle the uncertainty present in offline problems.

In portfolio optimization theory, investors demand a reward for bearing risk when making investment decisions. 
The economist Harry Markowitz introduced this risk-expected return relationship in the mean-variance model in a 1952 essay~\citep{r:6}, for which he was later awarded a Nobel Memorial Prize in Economic Sciences. The mean-variance model weighs the risk against the reward and solves the portfolio-selection problem (formally defined in Sec.~\ref{subsec:portfolio-opt}), i.e., decides how much wealth to invest in each asset by considering only the expected return and risk, expressed as the variance.

Our main contribution is a new risk-aware planning algorithm for offline RL  that mitigates the distributional shift problem, which is inspired by modern portfolio theory (MPT).
Our planning algorithm is s best-first search algorithm that treats the value associated with each candidate trajectory as an asset with specified mean and variance, which acts as a proxy for the risk indicating a distribution shift.
When building the search tree, we treat the limited node-expansion budget as wealth to invest in many different assets (each asset corresponds to a candidate trajectory) and solve a portfolio-optimization problem to determine how much wealth to invest in each of these assets. 
We refer to our planner, which we formally introduce in Sec.~\ref{sec:method}, as Wall Street Tree Search (\wsts) as an homage to the famous Monte Carlo Tree Search (\mcts)~\citep{r:50} algorithm and evaluate it in continuous control tasks from the widely studied D4RL offline RL benchmark~\citep{r:9}. 
We show, in Sec.~\ref{sec:result},  that \wsts{} matches or exceeds the performance of state-of-the-art (SOTA) offline RL algorithms. At the same time, it is substantially more reliable than the conventional decoding method resulting in significantly reduced variance when used on the same environment model. 

\section{Related Work}
Algorithms for offline RL typically fall under two categories: model-free and model-based algorithms.
Generally, each category addresses the uncertainty 
(mainly due to the distributional shift) differently. 
In \emph{model-free} algorithms, the agent learns a policy or value function directly from the dataset. Such algorithms typically address the distribution shift by constraining the learned policy to avoid out-of-distribution behavior that the dataset does not support~\citep{r:17, r:21, r:22, r:82, r:67} or use uncertainty quantification techniques, such as ensembles, to stabilize Q-functions~\citep{r:80, r:17, r:81, r:82}.

The second category,model-based RL methods, is the approach we take in this work, which is less explored for offline RL. Nonetheless, prior works have demonstrated promising results of MBRL methods, particularly in offline RL~\citep{r:24, r:25, r:3, r:4}. However, MBRL methods are subjected to the so-called model-exploitation problem, which causes the decision maker (the planner) to choose erroneous predictions with higher returns than would be obtained from the actual environment. 
\citet{r:92} addressed the model-exploitation problem by using ensembles and averaging the reward over all ensemble members.
\citet{r:24} addressed this problem by incorporating pessimism into learned dynamics models. 
\citet{r:25} also studied the effects of uncertainty in a model-based approach to offline RL and suggested optimizing a policy using an uncertainty-penalized reward. In contrast, we propose an explicit planning algorithm to account for the uncertainty.

Our approach to offline RL builds on the recent works by~\citet{r:3, r:4} that frame RL as a sequence-modeling problem and exploit the toolbox of contemporary sequence modeling.
At the heart of their approach is a sequence model based on the GPT-3 Transformer decoder architecture~\citep{r:5}, which they use for learning a distribution over trajectories by jointly modeling the states and actions to provide a bias toward generating in-distribution actions. 
To plan using the Transformer model (decode its outputs), \citeauthor{r:3} 
coupled it with a minimally modified version of beam search (\bs)~\citep{r:60}, and replaced the log probabilities of transitions with the expectation of the predicted reward signal to create a search strategy for reward-maximizing behavior.
However, decoding methods accounting only for the trajectories with the maximum predicted rewards and ignoring the uncertainty may suffice for large-scale language models, 
but might not be the best option in an offline RL setting.
We incorporate uncertainty into sequential decision-making to address this issue and propose a new decoding algorithm.

\section{Problem Definition}
\label{sec:preliminaries}
A \emph{Markov decision process} (MDP) is  a tuple  $\mathcal{M} = \{S, A, r, P, \rho_0, \gamma\}$
where 
$S$ and $A$ are the sets of states and actions,
$r : S \times A \rightarrow \mathbb{R}$ is the reward function,
$P$ is the system's dynamics, which is a conditional probability distribution of the form $P(s_{t+1}|s_t, a_t)$, representing the probability over the next state given the current state and the applied action,~$\rho_0$ defines the initial state distribution
and $\gamma \in (0, 1]$ is a scalar discount factor that penalizes future rewards.

A \emph{trajectory} $\upsilon = (s_1, a_1, r_1, s_2, a_2, r_2, . . . , s_{H} , a_{H} , r_{H})$ is a sequence of~$H$ states, actions and rewards. 
Here both the reward and the next state of a given state-action pair adhere to some underlying MDP.
A \emph{policy} $\pi: S \rightarrow A$ is a mapping between states and actions.

In the \emph{offline RL} problem, we are given a static, previously-collected dataset $\mathcal{D}$ of trajectories collected from some (potentially unknown) behavior policy~$\pi_B$ (for example,~$\pi_{B}$ can be human demonstrations, random robot explorations, or both).
The goal is to use $\mathcal{D}$ to learn, in an offline phase, a policy~$\pi$ for the underlying, unknown MDP $\mathcal{M}$, to be executed in an online phase.

\section{Algorithmic Background}
\label{sec:alg_background}
\begin{figure*}[t]
\centering
\includegraphics[width=17cm]{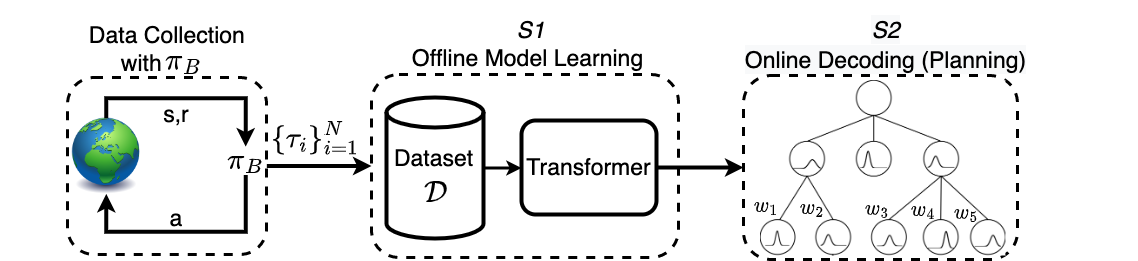} 
\caption{\label{fig:model}
\emph{Data collection:} The behavior policy $\pi_B$ collects the training batch $\mathcal{D}$.
\emph{Model Learning (S1, offline):} Given the previously collected batch, $\mathcal{D}$, we learn a sequence model.
\emph{Decoding (S2, online):} Decides which partial trajectories to simulate at each time point based on some scoring function outputs: $w_1, ..., w_N$ (we formally define {\tt \textbf{score}} in \ref{subsec:decoding})
}
\label{fig:top_level}
\end{figure*}

We take a \emph{model-based} offline RL approach in this work. Such an approach divides the problem into two stages
\textbf{S1}:~learning an MDP model $\mathcal{M}'$ that approximates $\mathcal{M}$ using the dataset $\mathcal{D}$ and  \textbf{S2}:~using the learned model $\mathcal{M}'$ to extract the policy~$\pi$.
We will refer to the first and second stages as the \emph{offline model learning} and the \emph{online decoding} stages, respectively, and to the entire approach as \emph{model-based RL}.
We note that we use the same Transformer architecture proposed by ~\citet{r:3} to learn an MDP model~\textbf{(S1)}, but change the decoding procedure~\textbf{(S2)}.
This overall approach is visualized in Fig.~\ref{fig:top_level}.
As stated, our work focuses on the decoding procedure~\textbf{(S2)}, the right block in Fig.~\ref{fig:top_level}.

\subsection{Offline Model Learning (S1)} \label{subsec:model_learning}
As previously mentioned, our approach to offline RL builds heavily on recent model-based offline RL methods~\citep{r:3, r:4} using Transformers to learn an MDP model~\textbf{(S1)}. We describe this approach in this subsection.

\subsubsection {Dataset} \label{subsec:dataset}
To create the dataset used to train the Transformer model, each transition in the original trajectory~$\upsilon$ (Sec. \ref{sec:preliminaries}) is augmented with a discounted reward-to-go estimate (namely,~$R_t = \sum_{t' = t}^H \gamma^{t'-t}r_t'$) to obtain an \emph{augmented trajectory} consisting of a sequence of~$H$ states, actions, rewards, and reward-to-go.
Here, it is important to note that the estimated reward-to-go is computed using the training data and estimates the return obtained by following the behavior policy~$\pi_B$. 
In general, it does not necessarily approximate the values of the learned policy. However, since we use it as a heuristic estimation, it suffices for guiding the search in the decoding stage~\citep{r:3}.

Using a discrete-token architecture (a Transformer) forces tokenization of the augmented trajectory. This is done by discretizing each dimension independently in the event of continuous inputs.
Assuming~$N$-dimensional states,~$M$-dimensional actions, and scalar reward and reward-to-go, the \emph{tokenized trajectory}~$\mathbf{y}$ is a sequence of length~$T=H\cdot(N +M +2)$, where every dimension of the trajectory is a token (subscripts on all tokens denote timestep, and superscripts on states and actions denote dimension.): 
$$
\mathbf{y}
= (\ldots, 
s_t^1, s_t^2,\ldots, s_t^N, 
a_t^1, a_t^2,\ldots,a_t^M,  
r_t, R_t,
s_{t+1}^1
\ldots).
$$
\subsubsection {Notation}
The elements of the vector~$\mathbf{y}$ (a tokenized trajectory) are denoted using an ordered set of components, $y_t$, $\mathbf{y}=(y_1, y_2,\ldots, y_T)$.
Each $y_t$ is an element of $\mathcal{V}$, the set of output tokens.
 $\mathbf{y}_{<t}$ denotes a trajectory from the first time step up to $t-1$,   
$\mathcal{Y}$ is the set of all valid output trajectories,
$x$ is the initial state, sampled from~$\rho_0$ (the initial state distribution of the MDP $\mathcal{M}$). Finally, $\circ$ denotes a concatenation operator.

\subsubsection {Model} \label{subsubsec:model}
A \emph{Transformer}~\citep{r:57} is a sequence-transduction model whose network architecture relies solely on attention mechanisms~\citep{r:58}, which allows for high parallelization and is shown to be highly effective in natural language processing (NLP) tasks~\citep{r:12, r:5}.
In our case, we use the Transformer to select an action sequence to be executed in the environment.

The model (parametrized via the network's weights $\theta$) predicts the probability distribution over an output space of possible trajectories~$\mathbf{y}$ given an initial state $x$, which is factorized as:
$$P_{\theta}(\mathbf{y}|x)=\prod_{t=1}^T P_{\theta}(y_t| \mathbf{y}_{<t}, x).$$
Where $P_{\theta}(y_t| \mathbf{y}_{<t}, x)$ 
typically defines a multinomial classification model. In such case, assuming a per-dimension vocabulary size of $\mathcal{V}$, the network's output layer consists of logits over a vocabulary of size $\mathcal{V}$. 
Namely, for a discrete output variable $y_t$ (which represents a dimension of a state, action, reward, or reward-to-go estimate), the outputs correspond to a mapping: $y_t^i\mapsto p_i$ for $i \in \{1, \ldots \mathcal{V}\}$ where $p_i$ is the probability of the next token to be $y_t^i$.
 
As we use the original network as-is, we refer the reader to the original paper for the architecture's details.

\subsection{Online Decoding (S2)}\label{subsec:decoding}
To decode a sequence from a trained model, we use the model's autoregressive nature to predict a single token, $y_t$, at each step.
Given an initial state $x$, we generate tokens sequentially using some heuristic-based algorithms.
The core algorithm providing the foundation of our planning techniques is beam search (\bs)~\citep{r:60}. 
Here we describe \bs{} as a meta-algorithm. As we will see in Sec.~\ref{sec:method},  this will significantly simplify the presentation of our planning algorithm \wsts{}. 
Accordingly, we present in Alg.~\ref{alg:beam_search} \bs{} as an algorithm that takes in additional two functions as inputs- a scoring function {\tt \textbf{score}} and a filter function {\tt \textbf{filter}} and returns a single approximated \textit{best} trajectory (we will define \textit{best} shortly). 
At each iteration, \bs{} expands all currently-considered sequences and creates new candidate sequences. These candidate sequences are evaluated using the {\tt \textbf{score}} function. Subsequently, \bs{} uses these scores as inputs to the {\tt \textbf{filter}} function and decides which~$\mathcal{B}$ sequences to keep for the next iteration.

\begin{algorithm}[tb]
\caption{Beam Search (BS)}
\label{alg:beam_search}
\textbf{Input}: start state $x$, scoring function {\tt \textbf{score()}}, filter function {\tt \textbf{filter()}}, sequence model $P_{\theta}(y_t| \mathbf{y}_{<t}, x)$ \\
\textbf{Parameters}: beam width $\mathcal{B}$, planning horizon $H$ \\
\textbf{Output}: Approx. single best trajectory.

\begin{algorithmic} [0] 

\STATE $C_0 \gets \{x\} $

\FOR{$t \gets 1$ to $H$}   
\STATE $C \gets \{\} $
\FORALL{$ \mathbf{y}_{<t} \in C_{t-1}$}

\STATE \texttt{// Autoregressively simulate transitions}
\STATE $(s_t, a_t, r_t, R _t) \sim  P_{\theta}(y_t| \mathbf{y}_{<t}, x)$
\STATE $C\leftarrow C \cup  \{ (\mathbf{y}_{<t} \circ (s_t, a_t, r_t, R _t)) \}$
\ENDFOR
\STATE $\mathbf{w} \gets {\tt \textbf{score}}(C)$
\STATE $C_t \gets {\tt \textbf{filter}}(C, \mathcal{B},\mathbf{w})$
\ENDFOR
\STATE \textbf{return} $C.max()$
\end{algorithmic}
\end{algorithm}

As we'll see next, different definitions of "best" result in different implementations of these functions and, in turn, correspond to different search algorithms that may be used during online decoding.
For example, in NLP and imitation learning, the overarching objective is to find the most probable sequence under the model at inference times (commonly known as maximum a posteriori, or MAP, decoding~\citep{r:8}).
This corresponds to solving the following optimization problem:
\begin{equation} 
\label{eq:MAP}
    y^{*} = {\arg \max}_{\mathbf{y} \in \mathcal{Y}} \log  P_{\theta}(\mathbf{y}|x).
\end{equation}
In this case, finding an exact solution is computationally hard. An approximate solution is often found using \textsf{top-$K$}~sampling~\citep{r:16}, a widely used \bs{} variant with the following selection of scoring and filter functions: 
The scoring function is the conditional probability distribution ${\tt score}(\mathbf{y}_{<t} \circ y_t) = \log P_{\theta}(y_t|\mathbf{y}_{<t})$. 
The filter function randomly picks the $\mathcal{B}$ most likely next tokens with proportion to their score.

In the context of offline RL, \citet{r:3} used a different {\tt \textbf{score()}} function and modified the top-K sampling \bs{} variant to decode trajectories that achieve the maximum cumulative expected rewards. 
For this case, the log probabilities of transitions are replaced by the log probability of the predicted reward signal. Consequently, the scoring function is the expected cumulative reward added to the reward-to-go estimate while the filter function remains unchanged. 
For simplicity, here we will refer to this variant simply as Expectation Maximizing Beam Search \rmbs.

As we will see shortly (Sec.~\ref{sec:method}), 
we explain \wsts{} using a particular choice of these functions.

\subsection{Portfolio Optimization}
\label{subsec:portfolio-opt}
In finance, a \emph{portfolio} is defined as a combination of financial assets, each typically associated with some expected reward and risk. To form the portfolio, given such $N$ different risky assets and wealth $w$, we need to decide how much wealth to invest in each asset, i.e., determine the vector of weights on assets $1$ to $N$: $w = (w_1,w_2, ..., w_N)$.
The \emph{portfolio optimization} problem corresponds to selecting the best portfolio out of the set of all portfolios being considered, according to some objective. 
These objectives typically balance expected reward and risk allowing investors a principled way to maximize return while bounding risk.
Modern portfolio theory~\citep{r:6} or mean-variance analysis solves the portfolio selection problem by taking only the mean and variance of the portfolio into consideration, weighing the risk, expressed as variance, against the expected return.

\emph{Expected utility theory} estimates the utility of action when the outcome is uncertain. It takes into account that individuals may be risk-averse, meaning they tend to prefer outcomes with low uncertainty to those with high uncertainty~\citep{r:56}. 
The expected utility maximization is consistent with mean-variance analysis if the utility function is quadratic or if the asset returns are normally distributed~\citep{r:55}. In such a case, the constrained portfolio-optimization problem can be formulated as a utility-maximization problem.
Specifically, given
some risk-aversion parameter~$\delta$,
assets mean and variance vectors~$\mu$ and $\Sigma$,
the  portfolio-optimization problem  is solved by computing a weight vector $w$ dictating the relative amount to invest in each asset by solving the following optimization problem:
\begin{equation}
    \label{eq:portfolio_opt}
    \max_w \quad w^T \mu - \frac{\delta}{2}w^T \Sigma w.
\end{equation}

\section{Method: Wall Street Tree Search (\wsts)}
\label{sec:method}
In this section, we present ``Wall Street Tree Search'' \:(\wsts)---our approach for online risk-aware decoding.
\wsts{} expands beam search with portfolio optimization to create a new planning algorithm for sequential decision-making under uncertainty. Like portfolio optimization, our risk-aware planning algorithm is about budget allocation to different assets. However, in our setting, the budget is not money--- it is the computational effort, and the assets are the candidate trajectories.  
In our variation of beam search, we allocate our limited amount of node-expansion budget to determine which $\mathcal{B}$ trajectories we keep at each time step. To this end, we use portfolio optimization to weigh trajectories according to their ``risk'' (which is uncertainty in our setting), expected return, and our risk-aversion parameter~$\delta$ (we discuss the implications of this parameter in Sec.~\ref{sec:result}).

We start by describing how we compute the mean vector and covariance matrix that will be used in Eq.~\ref{eq:portfolio_opt}.
We then continue to detail how \wsts{} uses portfolio optimization to instantiate the generic {\tt \textbf{score}} and {\tt \textbf{filter}} functions described in Sec.~\ref{subsec:decoding}.

\subsubsection{Mean Vector and Covariance Matrix} \label{subsec:mean_cov}
Recall that to solve the portfolio-optimization problem (Eq.~\ref{eq:portfolio_opt}), in addition to the risk-aversion paramaeter $\delta$, we require
(i)~a vector of expected returns,~\(\mu\), and 
(ii)~a covariance matrix~\(\Sigma{}\) of the assets.
To estimate these quantities, we use the Transformer decoder sequence model trained in the model-learning stage (Sec.~\ref{subsec:model_learning}).
Namely, we use the variance computed from the Transformer's output as a measure of predictive uncertainty and as a proxy indicating the amount of distribution shift.
This approach is based on~\citet{r:45}, which showed that Transformer-based models are well-calibrated when trained with temperature scaling~\citep{r:46} as is done in our setting.

To start, recall that the Transformer is a multinomial classification model which at inference time autoregressively outputs a probability~$p_i$ of the subsequent trajectory token~$y_t^i$ at step~$t$, conditioned on the preceding trajectory tokens~$\mathbf{y}_{<t}$.
Hence, by conditioning on each of the partial trajectories, the mean $E[y_t]$ and variance $\mathrm{Var}[y_t]$ of each output random variable~$y_t$ are simply
$E[y_t] = \sum _{i=1}^{\mathcal{V}}p_{i}y_t^{i}$
and
$\mathrm{Var}[y_t] = \sum_{i=1}^{\mathcal{V}}p_{i}\cdot (y_t^{i}-\mu )^{2}$, respectively.

Now, using these values to predict the vector of expected returns,~$\mu$, and the covariance matrix~$\Sigma{}$ for all the candidate solutions on each time step requires more care.
The longer the effective planning horizon, the more error the model inaccuracy introduces.
Thus, in contrast to the common approach in RL where immediate rewards are incentivized over long-term rewards via a discount factor~$\gamma \in (0,1]$ which exponentially scales down the rewards after each step, in our case, we want to account for the potential effect of compounding future risks.
Specifically, in addition to scaling down the mean, we also scale up the variance, informing our downstream planner that the uncertainty increases for future predictions.
Consequently, 
the means vector,~\(\mu\), is an $N$-dimensional vector consisting of the discounted means of the cumulative reward plus the reward-to-go estimate associated with each one of the $N$ candidate trajectories. 
The covariance matrix,~\(\Sigma\), is an $N \times N$ diagonal matrix consisting of the discounted variances of the cumulative reward plus reward-to-go estimate associated with each candidate trajectory along its main diagonal.
Namely,~$\mu_j$ and~$\sigma_j$ which are the $j$th entry in~\(\mu\) and~\(\Sigma\), respectively are defined as:
\begin{align*}
    \mu_j   &=\sum_{t=1}^{T-1} \gamma^t \mathrm{E[r_t^i]}+ \gamma^T \mathrm{E[\hat{R}_{T}^i]},\\
    \sigma^2_j&=\sum_{t=1}^{T-1} \gamma^{-2t} \mathrm{Var[r_t]} + \gamma^{-2T} \mathrm{Var[\hat{R}_{T}]}.
\end{align*}

\subsubsection{{\tt \textbf{score}}} function:
Given  the mean vector~$\mu$, and the covariance matrix~$\Sigma$,
we compute a score for each of the $N$ candidate trajectories available at time step $t$ by solving a portfolio-optimization problem via Eq.~\ref{eq:portfolio_opt} with the input risk-aversion parameter,~$\delta$. 
The output of the {\tt \textbf{score}} function is the solution to the portfolio-optimization problem, namely the wealth coefficients $w_1, \ldots, w_N$, representing the proportion of the computational time (money) to invest in exploring each trajectory (asset).

\subsubsection{{\tt \textbf{filter}}} function:
Given the coefficients $w_1, \ldots, w_N$, which are computed using the above score function, our filter function samples~$\mathcal{B}$ trajectories (with repetition) where the $j$th trajectory is sampled with probability $w_j$.
We underline that a trajectory may be sampled more than once. In such cases, the search is biased toward more-promising trajectories. 
Thus, our practical beam width is smaller than~$\mathcal{B}$ (but never bigger).

\section{Experiments and Results}
\label{sec:result}
In our experiments, we aim to study the following questions: 
\begin{itemize}
    \item[\textbf{Q1}] How does \wsts \: perform compared to prior approaches?
    \item[\textbf{Q2}] How to choose the risk-aversion parameter $\delta$?
\end{itemize}

\subsubsection{Experimental setup:}
\begin{figure}[t]
\centering
\includegraphics[width=8cm]{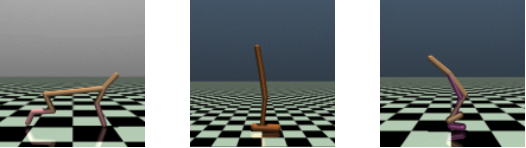} 
\caption{
The Gym-MuJoCo tasks, from left to right: HalfCheetah, Hopper,
and Walker2d.
}
\label{fig:mujoco}
\end{figure}

Our experimental evaluation focuses on continuous control tasks from the
MuJoCo physics engine, which are formulated as RL tasks and aims to facilitate research and development in robotics.
We focus on  Gym-MuJoCo locomotion tasks: Walker2d, HalfCheetah, and Hopper (displayed in Fig. \ref{fig:mujoco}).
To this end, we use the offline dataset from the widely used D4RL benchmark~\citep{r:9}.
The different dataset settings of D4RL are described below:

\begin{enumerate}
\item The ``medium'' dataset is generated by collecting 1 million samples from a partially-trained policy.
\item The ``medium-replay'' dataset records all samples observed during training until the policy reaches the ``medium'' level of performance.
\item The ``medium-expert'' dataset is generated by mixing 1 million samples generated by the medium policy concatenated with 1 million samples generated by an expert policy.
\end{enumerate}
We ran all tests on NVIDIA Tesla V100 GPUs.
For both planning algorithms, \rmbs{} and \wsts{}~\footnote{Code will be made publicly available upon paper publication}, we take an MPC approach, interleaving planning and execution (Sec.~\ref{sec:alg_background}). 
For both algorithms, we used the trained network weights and the same hyperparameters provided by the original paper's authors~\citep{r:3}.

\subsubsection{(Q1) How does \wsts \: perform compared to prior approaches?}
In this section, we compare \wsts{} with SOTA methods spanning
other approaches of offline RL. However, we focus our comparison on a \rmbs, which is the decoding algorithm employed in Trajectory Transformer~\citep{r:3}.
We concentrate our comparison on \rmbs{} for two reasons:

\begin{enumerate}
\item When using \rmbs{} to decode the Trajectory Transformer, it was shown to outperform prior offline RL methods on D4RL locomotion tasks.
We will shortly show that we have improved upon these results. Consequently, \wsts{} performs on par or better than the other offline RL methods (see Table \ref{table:results}). 
\item Our work focuses on the decoding procedure. Therefore we want to investigate the contribution of this individual component to the overall system performance. To this end, we modify only the overarching planning algorithm that decodes the Transformer's outputs from \rmbs{} to \wsts{}.
\end{enumerate}

\begin{table*}[t]
\centering
\begin{tabular}{l l l l l l l l l l}
\textbf{Dataset} &
\textbf{Environment}&
\textbf{CQL} &
\textbf{MOPO} &
\textbf{MBOP} &
\textbf{IQL} &
\textbf{DT} &
\textbf{TT (\rmbs{})} &
\textbf{WSTS} &
\textbf{$\delta$} \\
\hline

Med-Replay  &
HalfCheetah   &
$45.5 $  &
$\textbf{53.1} $  &
$42.3 $  &
$44.2 $  &
$36.6 $  &
$41.9 \pm 2.5$  &
$44.8 \pm 0.3$  &
$1.0$ \\
Med-Replay  &
Hopper &
$\textbf{95.0} $  &
$67.5 $  &
$12.4$  &
$94.7 $  &
$82.7 $  &
$91.5 \pm3 .6$  &
$94.2 \pm 2.6$  &
$2.0$ \\
Med-Replay  &
Walker2d  &
$77.2 $  &
$39.0 $  &
$9.7 $  &
$73.9 $  &
$66.6 $  &
$82.6 \pm 6.9$  &
$\textbf{86.1} \pm 3.8$  &
$2.0$ \\
\hline

Medium  &
HalfCheetah  &
$44.0 $  &
$42.3 $  &
$44.6$  &
$47.4 $  &
$42.6 $  &
$46.9 \pm 0.4$  &
$\textbf{47.6} \pm 0.2$ &
$0.1$ \\
Medium  &
Hopper &
$58.5 $  &
$28.0 $  &
$48.8$  &
$66.3 $  &
$\textbf{67.6} $  &
$67.4 \pm 2.9$  &
$67.0 \pm 4.7$ &
$0.5$ \\
Medium  &
Walker2d &
$72.5 $  &
$17.8 $  &
$41.0 $  &
$68.3 $  &
$74.0 $  &
$79.0 \pm 2.8$  &
$\textbf{82.1} \pm 0.8$ &
$1.0$ \\
\hline

Med-Expert  &
HalfCheetah  &
$91.6 $  &
$63.3$  &
$\textbf{105.0} $  &
$86.7 $  &
$86.8 $  &
$95.0 \pm 0.2$  &
$94.4 \pm 0.3 $ &
$0.1$ \\
Med-Expert  &
Hopper &
$105.4 $  &
$23.7 $  &
$55.1$  &
$91.5 $  &
$107.6 $  &
$110.0 \pm 2.7$  &
$\textbf{111.0}$ $\pm 1.6$  &
$0.1$ \\
Med-Expert  &
Walker2d &
$108.8 $  &
$44.6 $  &
$70.2 $  &
$\textbf{109.6} $  &
$108.1 $  &
$101.9 \pm 6.8$  &
$107.0 \pm 1.5$  &
$0.1$ \\
\hline

\textbf{Average}  &
 &
$77.6$  &
$42.14$  &
$47.8$  &
$75.8 $  &
$74.7$  &
$78.9$  &
$\textbf{81.6}$  \\
\end{tabular}
\caption{\wsts{} compared against other prior SOTA methods spanning other approaches of offline RL: 
\algName{CQL}~\citep{r:23}, 
\algName{MOPO}~\citep{r:25},  
\algName{MBOP}~\citep{r:92}, 
\algName{IQL}~\citep{r:95}, 
\algName{DT}~\citep{r:4} and Trajectory Transformer~(\algName{TT} which runs \rmbs{})~\citep{r:3}. variants correspond to the mean and standard error over 15 random seeds. Results for the other algorithms are taken from the original papers.}
\label{table:results}
\end{table*}

\begin{figure*}[t]
\centering
\includegraphics[width=15.9cm]{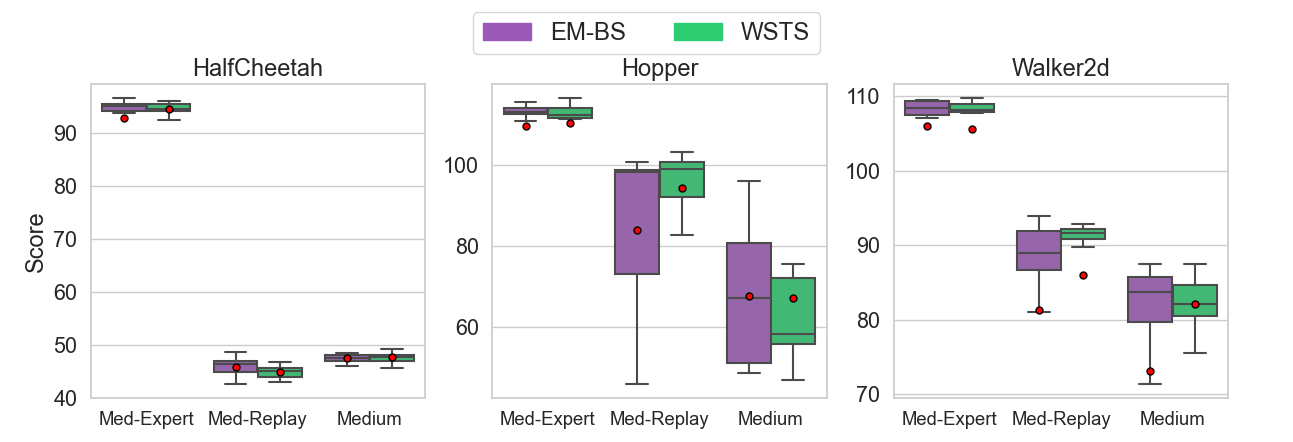} \caption{
Score as a function of the dataset for \wsts{} and \rmbs{} obtained over 15 random seeds.
The box corresponds to quartile~Q1 through Q3, with a line at the median (Q2) and a red dot on the mean. 
Whiskers are restricted to a maximum of 1.5 times the interquartile range. 
}
\label{fig:boxplot}
\end{figure*}

Table (\ref{table:results}) shows that \wsts{} outperforms \rmbs{}. However, reporting point estimates do not tell the whole story as they obscure some key aspects of the comparison. 
Indeed \wsts{} matches or surpasses the average score of \rmbs{}, but what is even more critical is that \wsts{} is consistently more stable, reducing the variance considerably.
To illustrate this, we first visualize the differences between \wsts{} and \rmbs{} using a box plot and present the results in Fig.~\ref{fig:boxplot}.

\begin{figure}[t]
\centering
\includegraphics[width=9.5cm]{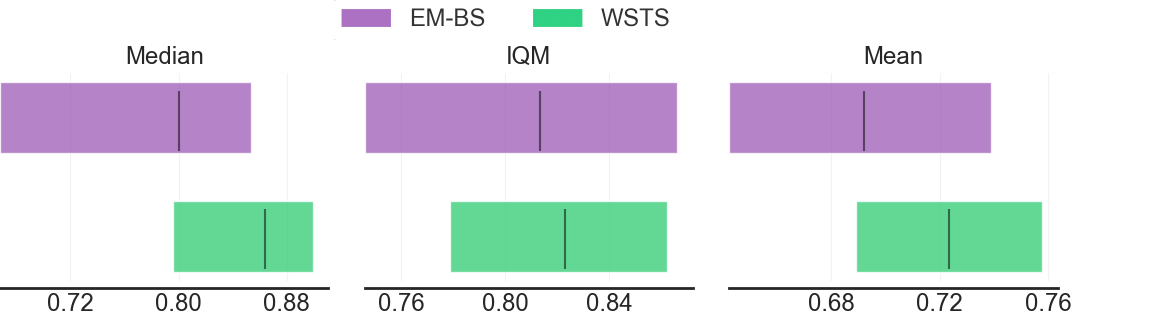} \caption{
Aggregate metrics on Gym-MuJoCo tasks with~$95 \%$ CI values based on 15 independent runs per each of the nine environments. 
The CIs are estimated using stratified sampling, with a total of $9 \times 15$ normalized scores per algorithm. The interval estimates are created with bootstrapping, i.e., calculating the aggregate metrics by randomly re-sampling scores with replacement from this data.
}
\label{fig:ci}
\end{figure}

Though the box plot helps understand better the difference between \rmbs{} and WSTS, it still doesn't provide a complete picture.   
Since evaluating runs is often computationally demanding in deep RL, only a small number of runs (usually less than 10) are generally assessed.
For such a low number of runs, reporting reliable results is challenging.
We address this by following the methodology proposed by~\citet{r:93}, which suggests presenting confidence intervals (CIs), which are a range of values that are likely to include the correct value with a certain degree of confidence.
Accordingly, in Fig. \ref{fig:ci},  we show $95 \%$ CI values of normalized mean, median, and interquartile mean (IQM) scores (performance on middle $50 \%$).
The CIs for \wsts{} are narrower and with higher averages than \rmbs{}, indicating that \wsts{} does not only outperform \rmbs{} but is also more stable than~\rmbs.

\subsubsection{(Q2) How to choose the risk aversion parameter $\delta$?}
Offline RL usually requires expensive online rollouts for hyperparameters search (for example, \citet{r:3} used six hyperparameters), which makes hyperparameters undesirable. 
However, in our case,  we can get a hint on how to set the risk aversion parameter's value properly by observing the connection between the risk aversion parameter and the offline training samples: scilicet, the training dataset size (relative to the environment complexity), and the data quality.

Generally, the risk-aversion parameter~$\delta$ takes the highest values for the medium-replay dataset; the medium dataset takes intermediate values; for the medium-expert dataset, the risk aversion parameter takes the lowest values, as seen in the rightmost column of Table~\ref{table:results}. 
This is expected as the more expert the behavior policy that generated the training samples is, the more we trust it and hence the less risk-averse we need to be.
This observation is concise with imitation learning. In the extreme case of expert demonstrations with unlimited data, the optimal strategy would be to select a single action with the maximum expected reward. Such an approach is equivalent to having $\delta-0$, which narrows the beam width $\mathcal{V}$ to one.

The number of training samples also plays a role when setting the value of $\delta$. More training samples reduce the subjective uncertainty, which in turn can reduce the risk aversion parameter value.
In our experiments, we can get affirmation for this observation when examining the comparison between \rmbs{} and \wsts{} in the medium-expert dataset (left boxplot in each image of Fig. \ref{fig:boxplot}).
Recall that a well-trained behavior policy generated the medium-expert dataset. In addition, this dataset contains twice as many samples as other datasets. For such a dataset, a behavior that tries to maximize the expected reward and does not account for the risk (i.e., low the risk-aversion parameter) is a good option. Indeed \wsts{} performs approximately as \rmbs{} does on this dataset.

\begin{figure}[t]
\centering
\includegraphics[width=7cm]{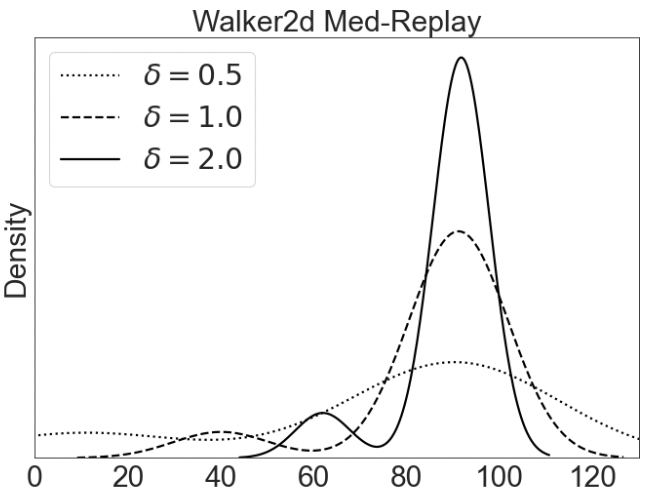}
\caption{\label{fig:parameter_tune} A kernel density estimate plot, presenting \wsts{} sensitivity to the risk aversion parameter $\delta$. Results
showed for Walker2d medium-replay dataset over ten random seeds for each risk aversion value.}
\end{figure}

It is also interesting to examine how the risk aversion parameter $\delta$ affects the performance. This question is relevant for scenarios where system engineers can select the amount of risk aversion. Such a selection generally depends on the application and system domain. Fig. \ref{fig:parameter_tune} presents a kernel density estimate plot, visualizing the distribution of scores in the Walker2d medium-replay dataset, that demonstrates that:
(i)~Our algorithm is robust when using diversified $\delta$ values
and that
(ii)~Decreasing the value of~$\delta$ allows achieving higher results if we are willing to take the risk of failing.

\section{Conclusion and Future Work}
\label{sec:conclusion}
The bitter lesson~\citep{r:96} is based on the historical observations that generic models tend to overtake specialized domain-specific approaches. 
This observation is manifested in a recently emerging body of work that shows that Transformer models that work well in other domains, such as natural language processing, can also provide an effective and generic solution to the offline RL problem. 
However, prediction with Transformers is currently made by repurposing the same tools from NLP, ignoring uncertainty. 
In this work, we addressed the question of \textit{how we can modify the decoding algorithms of large-scale language models-- the Transformer to tackle the uncertainty present in offline problems.}

To this end, we introduced \wsts, a planning algorithm that suggests a constructed way to control the risk.
Our planning algorithm leverages the capacity of the learned model to generalize to states outside the static batch support. Still, it is cautious when drifting to states where the model can't give a confident prediction based on the offline dataset.
For such states, our planner weighs this risk vs. the expected return by incorporating modern portfolio theory into sequential decision-making.

In this work,  we rely on temperature scaling for calibration.
However, a question remains whether different~Bayesian formulations of deep learning that can potentially improve predictive uncertainty quantification can further improve our results. 
Moreover, our planning algorithm penalizes trajectories with uncertainty. However, we are more interested in the risk of incurred loss than the variability of returns measured by standard deviation; hence we should only penalize for the risk of low returns (but we need not penalize the risk of high returns). We hope to explore these limitations in future work using better-suited portfolio optimization methods.

We demonstrated our algorithm on an offline model-based RL setting.
Our experiments show that our method for controlling the risk is advantageous in the offline RL setting, as \wsts{} outperforms state-of-the-art offline RL methods in the D4RL standard benchmark.
Moreover, it is consistently more stable, reducing the variance of the results considerably, an essential property in a range of real-world domains where real-world failures can be costly or dangerous.

We conclude with a quote by Warren Buffett~\citep{r:61}, representing his view on investors' uncertainty which also captures the essence of our approach to planning with an uncertain model applied to offline RL.
\begin{quote}
    There is nothing wrong with a `know nothing' investor who realizes it. The problem is when you are a `know nothing' investor, but you think you know something.
\end{quote}

\medskip
\small

\bibliography{elbaz.bib}
\clearpage
\end{document}